%% file: acl2023.tex
\pdfoutput=1

\documentclass[11pt]{article}

\usepackage{ACL2023}
\usepackage{colortbl}
\usepackage{amsmath} 
\usepackage{times}
\usepackage{latexsym}
\usepackage{multirow}
\usepackage[T1]{fontenc}

\usepackage[utf8]{inputenc}
\usepackage{colortbl}

\usepackage{microtype}
\usepackage{booktabs}
\usepackage{graphicx}
\usepackage{inconsolata}
\usepackage{xspace}
\usepackage{xcolor}
\usepackage{multirow}
\usepackage[normalem]{ulem}
\useunder{\uline}{\ul}{}

\newcommand\benchname{XL$^2$Bench\xspace}

\title{\benchname: A Benchmark for Extremely Long Context Understanding with Long-range Dependencies}

\author{Xuanfan Ni$^{1,2}$, Hengyi Cai$^{3}$, Xiaochi Wei$^4$, Shuaiqiang Wang$^4$, \\ \textbf{Dawei Yin}$^{4}$, \textbf{Piji Li}$^{1,2}$\thanks{$^*$Corresponding author.\newline 
\hspace*{0.53cm} $^{\text{1}}$Code and Data: \href{https://github.com/nuaa-nlp/XL2Bench}{https://github.com/nuaa-nlp/XL2Bench}} \\ $^1$Nanjing University of Aeronautics and Astronautics, Nanjing, China
\\ $^2$MIIT Key Laboratory of Pattern Analysis and Machine Intelligence, Nanjing, China
\\ $^3$Institute of Computing Technology, CAS, Beijing, China, \  $^4$Baidu Inc., Beijing, China
\\{\tt \{xuanfanni,pjli\}@nuaa.edu.cn}, \ \tt caihengyi@ict.ac.cn
\\ \tt \{weixiaochi, wangshuaiqiang\}@baidu.com, \  \tt yindawei@acm.org} 

\begin{document}
\maketitle
\begin{abstract}
Large Language Models (LLMs) have demonstrated remarkable performance across diverse tasks but are constrained by their small context window sizes. 
Various efforts have been proposed to expand the context window to accommodate even up to 200K input tokens.
Meanwhile, building high-quality benchmarks with much longer text lengths and more demanding tasks to provide comprehensive evaluations is of immense practical interest to facilitate long context understanding research of LLMs.
However, prior benchmarks create datasets that ostensibly cater to long-text comprehension by expanding the input of traditional tasks, which falls short to exhibit the unique characteristics of long-text understanding, including long dependency tasks and longer text length compatible with modern LLMs' context window size.
In this paper, we introduce a benchmark for e\textbf{X}tremely \textbf{L}ong context understanding with \textbf{L}ong-range dependencies, \textbf{\benchname}, which includes three scenarios—{Fiction Reading}, {Paper Reading}, and {Law Reading}—and four tasks of increasing complexity: {Memory  Retrieval}, {Detailed Understanding}, {Overall Understanding}, and {Open-ended Generation}, covering 27 subtasks in English and Chinese. 
It has an average length of 100K+ words (English) and 200K+ characters (Chinese).  
Evaluating six leading LLMs on \benchname, we find that their performance significantly lags behind human levels. 
Moreover, the observed decline in performance across both the original and enhanced datasets underscores the efficacy of our approach to mitigating data contamination.$^{\text{1}}$
\end{abstract}

\input{sections/Intro}
\input{sections/RelatedWork}
\input{sections/Method}
\input{sections/Exp}

\input{sections/Conclusion}

\bibliography{anthology,custom}
\bibliographystyle{acl_natbib}

\appendix

\input{sections/A-Description}
\input{sections/A-template}
\input{sections/A-All_Results}

\end{document}

%% file: sections/Intro.tex
\begin{figure}[!t]
	\centering
	\includegraphics[width=\linewidth]{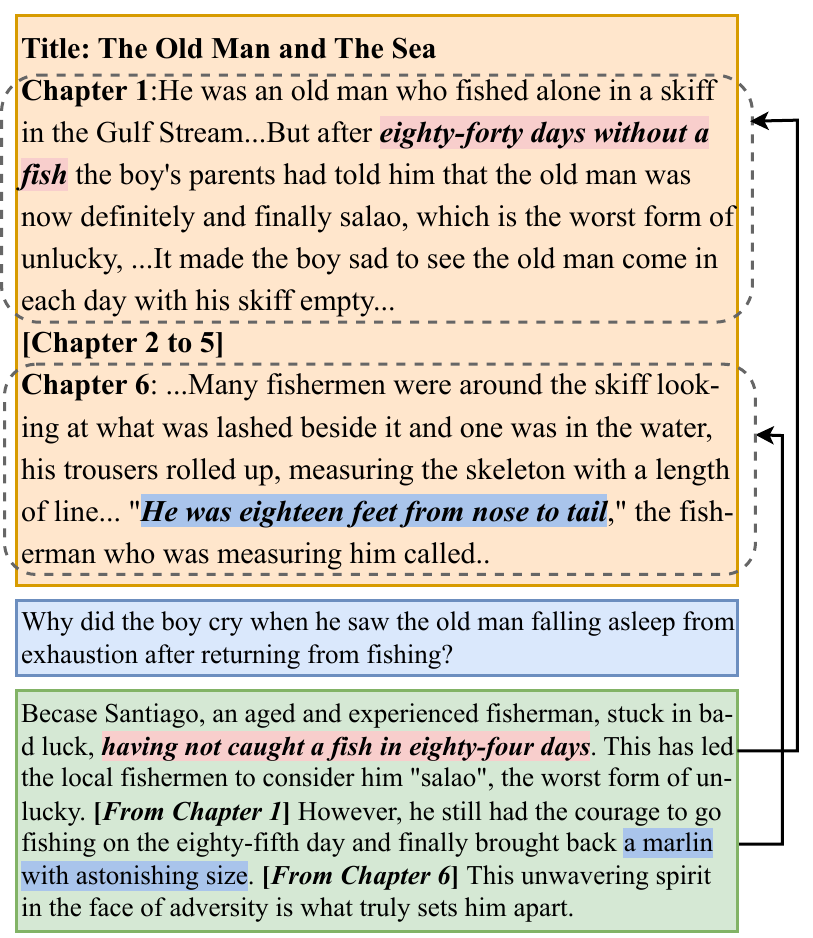}
	\caption{An illustrative example of long-dependency task, in which the model needs to make connective inferences across input document to fulfill the goal.}
	\label{example_x}
    \vspace{0mm}
\end{figure}

\section{Introduction}
Large Language Models (LLMs) have attracted considerable interest for their remarkable capabilities in a wide range of NLP tasks.
However, a common limitation among these models is the fixed context window size (for example, LLaMA with maximum 2048 tokens and GPT-3.5 with maximum 4096 tokens), rendering them incapable of memorizing and understanding extremely long inputs~\cite{DBLP:journals/corr/abs-2307-03172}.
Evidenced by a basic passkey retrieval task, the accuracy of LLaMA recalling a passkey plummets from nearly 100\% to nil when the text surpasses 2048 tokens~\cite{DBLP:journals/corr/abs-2307-03170}.

In pursuit of the goal of improving LLM's ability to comprehend long-context textual information, various efforts have been proposed to expand the context window of LLMs, such as sparse attention~\cite{DBLP:journals/corr/abs-2307-03170,DBLP:journals/corr/abs-2309-12307,DBLP:journals/corr/abs-2305-16300}, length extrapolation~\cite{DBLP:conf/acl/DaiYYCLS19, DBLP:journals/corr/abs-2104-09864,DBLP:journals/corr/abs-2309-00071}, and context compression~\cite{DBLP:journals/corr/abs-2307-06945,DBLP:journals/corr/abs-2304-08467}.
Given the notable advances achieved by these techniques, the necessity for high-quality benchmarks, featuring longer text lengths and more complex tasks, is escalating to facilitate thorough evaluations of LLMs' long context understanding ability.


Being able to understand long-range dependencies in context and be sensitive to various perturbations applied to distant context is what sets long text understanding apart from traditional NLP tasks~\cite{DBLP:journals/corr/abs-2009-06097,DBLP:conf/iclr/Tay0ASBPRYRM21,DBLP:conf/acl/RaeR20,DBLP:conf/emnlp/NiDRL23}.
Existing benchmarks for long-text understanding, such as LongBench~\cite{DBLP:journals/corr/abs-2308-14508}, L-Eval~\cite{DBLP:journals/corr/abs-2307-11088}, and InfiniteBench~\cite{zhang2023infinitebench}, often merely expand the input of traditional tasks to create datasets that ostensibly cater to long-text comprehension~\cite{DBLP:journals/corr/abs-2308-14508,DBLP:journals/corr/abs-2307-11088}. However, this approach does not tailor tasks to the distinct features of long-text comprehension, thereby impeding the thorough assessment of LLMs' abilities in understanding extended contexts.
Moreover, the average text length in existing benchmarks usually does not exceed a few thousand tokens, significantly shorter than the long texts perceived in human cognition.
For example, a user might upload an entire novel and inquire about the development of the protagonist's storyline. This task would require the model to process and comprehend texts spanning over ten thousands of words, necessitating long-range understanding and reasoning within the content to adequately address the question.
Traditional benchmarks typically fall short in measuring capabilities of LLMs to aggregate disparate pieces of information scattered throughout the whole input texts in more realistic scenarios, making it challenging to truly evaluate LLMs' ability on long context understanding~\cite{DBLP:journals/corr/abs-2309-13345,DBLP:journals/corr/abs-2310-19240}.

In light of the deficiencies identified in current benchmarks, this paper proposes a benchmark for e\textbf{X}tremely \textbf{L}ong context understanding with \textbf{L}ong-range dependencies, \textbf{\benchname},
which features three scenarios——{Fiction Reading}, {Paper Reading}, and {Law Reading}.
\benchname contains extremely long documents with an average of 100K+ words (English) and 200K+ characters (Chinese), along with 632K questions spanning over four specifically designed tasks to examine a model's ability to aggregate and compare information across long context, including  \textit{Memory Retrieval}, \textit{Detailed Understanding}, \textit{Overall Understanding}, and \textit{Open-ended Generation}.
These tasks mimic the way people use LLMs in real-world scenarios. 
Figure~\ref{example_x} illustrates a case where the model explains a boy's tears as stemming from a story about the old man who, against significant challenges, successfully captures a marlin. 
To construct a solid answer, it demands the model to identifies passages describing the boy's reaction, the man's triumph, and his earlier hardships across various chapters, and make connective inferences using details buried far back in the long context.

Besides, to address data contamination caused by outdated long texts contained in benchmark, we implement three data augmentation strategies: \textbf{text transformation}, which involves altering the original text into a different language or style; \textbf{text replacement}, which entails modifying or substituting key textual information; and \textbf{text concatenation}, which incorporates integrating additional texts into the original document.

Results of experiments on multiple state-of-the-art LLMs reveal that even the most advanced LLMs currently available fall short of reaching human-level proficiency on \benchname. Despite these models' ability to handle texts of considerable length, there is a marked decline in performance as the text lengthens. 
Additionally, the results obtained by RAG~\cite{DBLP:journals/corr/abs-2202-01110, DBLP:journals/corr/abs-2312-10997} on \benchname demonstrate that retrieval-based methods fail in overall and detailed understanding tasks; instead, they require that the models comprehensively grasp the entirety of the long texts. 
Furthermore, we conduct ablation experiments to compare model performance on both original and augmented benchmarks, which shows that the strategies we employ to address the issue of data contamination are indeed effective.


Our contributions are delineated as follows:
\begin{itemize}
    \item We construct \benchname, a comprehensive benchmark for extremely long text understanding with well-designed tasks.
    \item We formulate three data augmentation techniques to circumvent the issue of data contamination frequently encountered when using LLMs alongside existing NLP datasets. Through experimentation, we validate the efficacy of these methodologies in mitigating concerns about data contamination.
    \item We conduct empirical experiments to evaluate the performance of advanced LLMs using \benchname. The results reveal that contemporary LLMs are still facing challenges in achieving comprehensive understanding across long textual inputs.
\end{itemize}

%% file: sections/RelatedWork.tex
\section{Related Work}
\subsection{Long Context Modeling}
Large language models (LLMs), such as GPT-4~\cite{achiam2023gpt} and Llama~\cite{DBLP:journals/corr/abs-2302-13971,DBLP:journals/corr/abs-2307-09288}, have exhibited superior performance across a variety of text generation tasks and practical deployment scenarios~\cite{DBLP:journals/corr/abs-2312-03863,DBLP:journals/corr/abs-2310-19736,DBLP:conf/emnlp/WangLL23}. Nonetheless, the principal limitation hindering LLMs from harnessing their greater potential is the context window size---the upper limit of text length the model is capable of processing~\cite{DBLP:conf/acl/RatnerLBRMAKSLS23}. To circumvent this limitation, methods based on Position Encoding~\cite{DBLP:conf/naacl/ShawUV18}, length extrapolation~\cite{DBLP:conf/blackboxnlp/NewmanHLM20}, and sparse attention mechanisms~\cite{DBLP:conf/emnlp/ZhangTS21,DBLP:journals/ijufks/GaoL23}, such as Alibi~\cite{DBLP:conf/iclr/PressSL22,}, RoPE~\cite{DBLP:journals/corr/abs-2104-09864}, and Landmark~\cite{DBLP:journals/corr/abs-2305-16300}, have been presented.
Furthermore, some strategies compress texts to align with the model’s context window size~\cite{DBLP:journals/corr/abs-2304-08467,DBLP:conf/emnlp/ChevalierWAC23}.
Alternative approaches like Retrieval-Augmented Generation~\cite{cai2022recent,li2022survey} and Memory Bank~\cite{DBLP:journals/corr/abs-2306-07174} utilize segmented retrieval followed by generation.


\begin{table*}[!t]
\centering
\resizebox{2\columnwidth}{!}{
\begin{tabular}{lllrrrrr}
\toprule[1.3pt]
\textbf{Tasks}                       & \textbf{Subtasks}         & \textbf{Source}     & \multicolumn{2}{c}{\textbf{Num}}               &  \multicolumn{2}{c}{\textbf{Avg. Len}}        &\textbf{Metric}     \\  \cmidrule(lr){4-5} \cmidrule(lr){6-7}
& & &CN & EN &CN & EN & \\ \hline
\rowcolor{gray!50}
\multicolumn{8}{c}{\cellcolor{gray!50}\textit{\textbf{Fiction Reading}}}  \\ 

\multirow{2}{*}{Memory Retrieval}     & Content Location          & Content Extraction  &1495 &1405  & 571.6K          &  111.5K                  & Acc.     \\     
& Content Retrieval          & Content Extraction      &  299& 261 &571.1K &116.0K & Acc.         \\ \cmidrule{1-8}
\multirow{2}{*}{Detailed Understanding}  & Chapter Summarization & Data Synthesis & 167&156& 569.7K & 110.6K     & Rouge-L   \\ 
                  & Question Answering        & Data Synthesis     & 249 &269 &562.0K & 114.7K & BLEU \\  \cmidrule{1-8}
\multirow{6}{*}{Overall Understanding}   & Chapter Counting          & Content Extraction         & 30 &27&569.7K&113.4K&Acc. \\ 
    & Background Summarization             & Data Synthesis         & 30&27&570.3K & 113.7K  &Rouge-L   \\ 
         & Event Extraction                  & Data Synthesis         &30 &27&570.2K&113.7K & Rouge-L       \\ 
            & Fiction Summarization & Data Synthesis        &30 &27&570.4K&113.8K&Rouge-L            \\
   & Character Description              & Data Synthesis  & 191&140&589.7K& 143.5K & Rouge-L \\
   & Relationship Analysis              & Data Synthesis  &193&432&606.3K&189.8K&Rouge-L     \\ \cmidrule{1-8}
 \multirow{3}{*}{Open-ended Generation} & Role-play Conversation &Data Synthesis     & 293  &256 &592.7K &115.2K &BLEU         \\ 
                    & News Generation   & Data Synthesis     &30 &27 & 570.7K&114.0K & BLEU    \\ 
       & Poem Generation        & Data Synthesis         & 30&27&570.1K&113.6K &BLEU\\ \hline

\rowcolor{gray!50}
\multicolumn{8}{c}{\cellcolor{gray!50}\textit{\textbf{Paper Reading}}}  \\ 
  Memory Retrieval                      & Content Retrieval        & Content Extraction         &-&4532&-&13.7K&Acc.                        \\ \cmidrule{1-8}
 \multirow{2}{*}{Detailed Understanding}  & Section Summarization  & Data Synthesis         & -&3136&-&  14.1K & Rouge-L   \\ 
  & 
Terminology Explanation                & Data Synthesis    & -&14981&-&13.5K & BLEU                                                   \\ \cmidrule{1-8}
 \multirow{2}{*}{Overall Understanding}    & Paper Counting                  & Content Extraction         & - & 3100&-&13.5K &Acc.                                           \\ 
                        & Paper Summarization          & Data Integration         &-&518&-&14.0K&Rouge-L   \\ \cmidrule{1-8}
 \multirow{2}{*}{Open-ended Generation} & Paper Review                 & Data Integration   &-&518&-&14.0K & BLEU                                                                       \\
   & Rating Score                  & Data Integration         &-&518&-&13.6K &MAE     \\ \hline

\rowcolor{gray!50}
\multicolumn{8}{c}{\cellcolor{gray!50}\textit{\textbf{Law Reading}}}  \\ 
   \multirow{2}{*}{Memory Retrieval}     & Legal Entry Location              & Content Extraction        & 2213&-&105.6K&-&Acc.                                                 \\
                & Legal Entry Retrieval              & Content Extraction       &2225&-&105.3K&-&Acc.   \\ \cmidrule{1-8}
 \multirow{2}{*}{Detailed Understanding}                & Legal Definition QA    &   Data Synthesis &2635 &-&102.9K &-&BLEU \\
 & Legal Number QA & Data Synthesis & 1477&-&105.7K &- & Acc.\\  \cmidrule{1-8}
 \multirow{2}{*}{Overall Understanding}                     & 
Legal Entry Counting                 & Content Extraction& 122&-&103.0K&-&Acc.                                                                   \\
 &  Multiple Choice QA         &Data Integration     & 16881     &-       & 95.6K  &-&F1              \\ \cmidrule{1-8}
       Open-ended Generation           & Case Adjudication    &Data Integration            & 588369  &-&72.7K&-           & Acc.         \\ \bottomrule[1.3pt]                                                          
\end{tabular}
}
\caption{An overview of the statistics of \benchname. \textbf{Source} represents the method we use to construct the dataset for this subtask. \textbf{Num} represents the number of <input, output> pairs this subtask possesses. \textbf{Avg. Len} denotes the average combined length of the input and output, which is computed using the number of characters for Chinese and the number of words for English. \textbf{K} stands for 1024. For example, 200K = 200*1024.
\label{benchmark}}
\end{table*}

\begin{figure*}[!t]
   \centering
   \includegraphics[width=\linewidth]{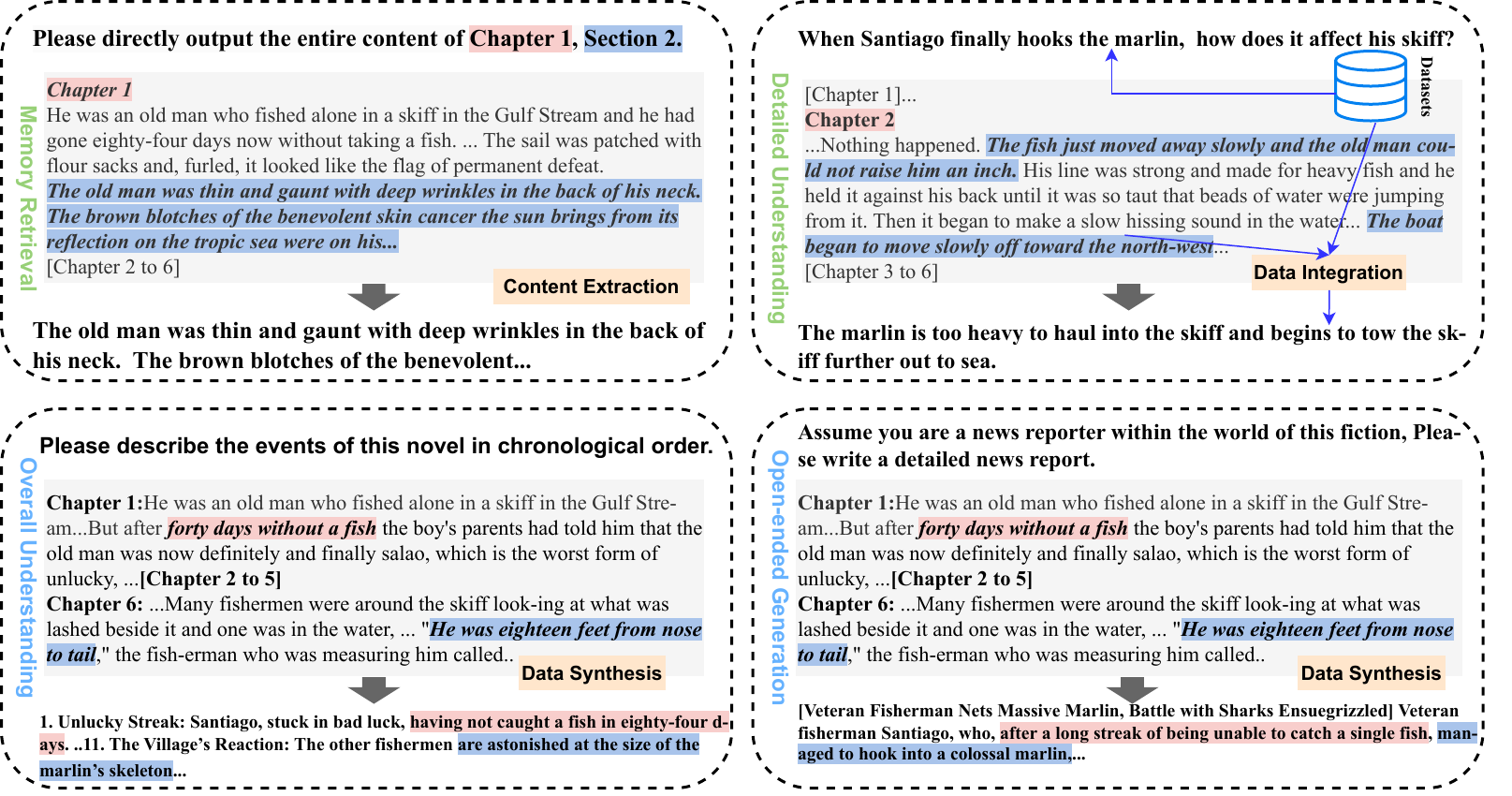}
        \caption{Illustration of the designed long context understanding tasks.}
    \label{framework}
    \vspace{-5mm}
\end{figure*}

\subsection{Evaluation Benchmarks}

Current research is frequently directed at developing benchmarks tailored to specific tasks, such as reasoning~\cite{DBLP:journals/corr/abs-2306-09212}, code~\cite{DBLP:journals/corr/abs-2107-03374,DBLP:journals/corr/abs-2108-07732}, role-play~\cite{wang2024characteristic}, and mathematics~\cite{DBLP:conf/nips/HendrycksBKABTS21,DBLP:journals/corr/abs-2110-14168,DBLP:journals/corr/abs-2305-12474}. However, existing benchmarks, such as LongBench~\cite{DBLP:journals/corr/abs-2308-14508}, L-Eval~\cite{DBLP:journals/corr/abs-2307-11088}, and Bamboo~\cite{DBLP:journals/corr/abs-2309-13345}, essentially expand existing NLU datasets, which may not pose sufficient difficulty and are prone to data contamination, and often fall short in text length. Besides, M4LE~\cite{DBLP:journals/corr/abs-2310-19240} offers control over text length within benchmarks. it constructs texts from fragments of multiple summarization datasets, which compromises textual cohesion. InfiniteBench~\cite{zhang2023infinitebench} introduces a broader range of tasks. However, the manual annotation required for such a benchmark is extremely costly. By way of contrast, \benchname leverages LLMs and meticulous human review to construct the benchmark cost-effectively. 

%% file: sections/Method.tex
\section{Methodology}
In this section, we introduce the construction methodologies of \benchname and design of tasks with various level of difficulty.





\subsection{Task Design}

We evaluate the model's understanding of extremely long texts from the perspectives of fine-grained retrieval and coarse-grained understanding. Based on this, we design four tasks: \textit{Memory Retrieval}, \textit{Detailed Understanding}, \textit{Overall Understanding}, and \textit{Open-ended Generation}.

\paragraph{Memory Retrieval.} This task challenges the model to accurately retrieve and respond to queries by finding content within the text that aligns with given instructions. For instance, the model may be asked to pinpoint the specifics of a legal entry within a law or identify the originating chapter of a passage from a novel, thereby evaluating its capability to accurately locate and interpret question-relevant content.

\paragraph{Detailed Understanding.} Here, the model is tasked with not only retrieving content but also comprehensively understanding it to perform activities such as summarization or question answering. This demands a more profound level of textual comprehension, surpassing mere content retrieval to include an in-depth analysis and synthesis of the text.

\paragraph{Overall Understanding.} To circumvent tasks being completed through simple content retrieval, we introduce the Overall Understanding task. This task necessitates a holistic comprehension of the long text, enabling the model to build long-range dependencies and tackle inquiries related to overarching themes, such as the depiction of a character throughout a novel or the trajectory of a company's stock across its history.

\paragraph{Open-ended Generation.} Building on a solid foundation of long text understanding, the model is expected to undertake generation tasks rooted in it, such as role-playing a character in the fiction. Outputs should demonstrate creative expansion and inference, adhering to the text's core themes and concepts, while ensuring originality and thematic consistency.

Table~\ref{benchmark} delineates the various subtasks encapsulated within these four primary tasks. For more task descriptions of \benchname, please refer to Appendix~\ref{detailed description}.

\subsection{Benchmark Construction}
In this subsection, we describe the sources from which we gather data and the methodologies we employ for constructing the benchmark in three different scenarios.

We gather long texts categorized under three scenarios. For fiction reading, we select a variety of novels written in both Chinese and English. For paper reading, we download PDF versions and reviews of papers submitted to ICLR 2023 from Openreview\footnote{https://openreview.net/group?id=ICLR.cc/2023/Conference}. For law reading, we gather a substantial collection of original Chinese legislations.

To minimize cost of human annotation, we employ three methods to construct : \textit{Content Extraction}, \textit{Data Integration,} and \textit{Data Synthesis}.

\paragraph{Content Extraction.} We extract content from the original text to serve as the answer and use the index of this portion of the content to formulate the question. For instance, we used the title of a paper as the answer, with the corresponding question being: \textit{What is the title of this paper?}

\paragraph{Data Integration.} Tasks within certain short text datasets bear formal resemblance to what we have designed, exemplified by Document QA. Consequently, we contemplate leveraging these datasets to augment our benchmark. More precisely, we employ LLMs to facilitate the alignment of data from the pre-existing datasets with our collected long texts. In an effort to mitigate potential overestimations of performance resulting from the model's familiarity with these datasets during its training phase, we utilize LLMs to meticulously rewrite the original texts and remove any information that may indicate the data source.

\paragraph{Data Synthesis.} For tasks that lack corresponding datasets, we utilize LLMs for direct generation. For summarization tasks, we employ structured text summarization method~\cite{chang2023booookscore} via LLMs. For QA tasks, we use in-context learning~\cite{brown2020language} to construct some examples for the model to generate.

Employing the aforementioned approaches, we have constructed an extremely-long text benchmark encompassing three distinct scenarios, four overarching tasks, 27 detailed subtasks, and a corpus of 700+ texts with a average length of 100K+ words for English and 200K+ characters for Chinese. The statistics of our benchmark are shown in Table~\ref{benchmark}.

\subsection{Human Verification} 
Regarding the use of LLMs to mitigate the costs associated with manual annotation, these approaches inherently limits our benchmark to the quality of the content produced by LLMs. However, it is important to \textbf{note} that: (1) XL$^2$Bench is not solely comprised of LLM-generated questions and answers, as these constitute no more than \textbf{30\%} of the benchmark. The majority of the content is derived from Content Extraction and Data Integration processes. (2) For the portion generated by LLMs, we implement a meticulous human verification process to ensure the quality of the questions and answers. This verification process involves:
\begin{itemize}
    \item We initially rule out content in the model’s response that is irrelevant to the text, such as phrases like ``\textit{Sure!}'', ``\textit{Of course!}'', ``\textit{Here are the answers.}'', etc.
    \item Next, we report inconsistencies between the report and the text, such as erroneous summaries of a different novel or the inclusion of a protagonist’s name from another novel in the summary.
    \item The model is then prompted to regenerate the content. If it still cannot produce the correct answers, human annotations are made. 
\end{itemize}
Through these operations, we strive to avoid the benchmark decline caused by LLMs generation.

\begin{table*}[!t]

\resizebox{2.1\columnwidth}{!}{
\begin{tabular}{l|cc|cc|cccccc|ccc}
\toprule[1.3pt]
\multirow{2}{*}{\textbf{Models}} & \multicolumn{2}{c|}{\textbf{MR}}                 & \multicolumn{2}{c|}{\textbf{DU}}          & \multicolumn{6}{c|}{\textbf{OU}}                                                                                        & \multicolumn{3}{c}{\textbf{TG}}      \\ \cmidrule(lr){2-3} \cmidrule(lr){4-5} \cmidrule(lr){6-11} \cmidrule(lr){12-14}  
                                 & \textit{C-L} & \textit{C-R} & \textit{C-S} & \textit{QA} & \textit{C-C} & \textit{B-S} & \textit{E-E} & \textit{F-S} & \textit{Ch-D} & \textit{Re-A} & \textit{RP-C} & \textit{N-G} & \textit{P-G} \\ \cmidrule{1-14}
YaRN-Mistral-7B            & \textless{}1           & \textless{}1           & 4.46                & 2.26               & 13.78              & 8.09               & 16.17             & 5.52                 & 8.35            & 7.91            & 7.28                 & 4.42            & 5.91            \\
InternLM2-C-7B         & \textless{}1           & \textless{}1           & 8.27                & \textless{}1       & 6.67               & 11.68              & 9.97              & 11.97                & 6.92            & 2.22            & 1.16                 & 5.88            & 3.49            \\
InternLM2-C-20B        & 6.85                   & \textless{}1           & 17.22               & 9.82               & 53.33              & 15.58              & 18.61             & 17.29                & 21.98           & 28.92           & 11.65                & 16.67           & 10.09           \\
Moonshot-V1                 & {\ul 60.39}            & {\ul 17.23}            & \textbf{23.53}      & {\ul 33.13}        & \textbf{86.30}     & \textbf{24.32}     & 20.08             & \textbf{25.10}       & {\ul 22.24}     & \textbf{54.99}  & 12.81                & {\ul 27.31}     & {\ul 12.22}     \\
GLM-4                       & \textbf{63.44}         & \textbf{20.08}         & 18.12               & 14.51              & {\ul 72.73}        & 18.40              & {\ul 20.42}       & 15.84                & 22.22           & 42.27           & {\ul 13.62}          & 19.70           & 11.69           \\
GPT-4-Turbo                & 54.36                  & 11.89                  & {\ul 19.87}         & \textbf{37.23}     & 60.00              & {\ul 21.21}        & \textbf{21.40}    & {\ul 21.57}          & \textbf{23.14}  & {\ul 49.05}     & \textbf{17.58}       & \textbf{30.19}  & \textbf{16.56}  \\ \bottomrule[1.3pt]
\end{tabular}
}
\caption{Results (\%) of six LLMs on Chinese Fiction Reading. \textbf{MR}, \textbf{DU}, \textbf{OU}, \textbf{TG} are the abbreviations for the initials of four tasks. \textbf{C-L}, \textbf{C-R}, \textbf{C-S}, etc., represent the abbreviations of 13 subtasks. The context window size of GLM-4 and InternLM2-Chat is 200K, whereas it is 128K for other models. The \textbf{bold} numbers in the results represent the best scores, whereas the \underline{underlined} numbers indicate the second-best scores.\label{results_fiction_cn} }
\end{table*}

\subsection{Data Contamination} 
The potential of data contamination warrants serious consideration when constructing a benchmark~\cite{DBLP:conf/emnlp/SainzCGELA23, DBLP:journals/corr/abs-2311-09783, DBLP:conf/acl/Magar022}. The risk arises when the test set data is either identical to, or strikingly similar to, the training set data. This could result in the model memorizing specific answers instead of acquiring the ability to reason or generalize from unseen data. In our construction process, the selected novels, academic papers, and legal texts may have been included in the training corpus of LLMs. Consequently, the model may not need to fully comprehend the entire text to accomplish various tasks. In order to mitigate the impact of data contamination on model's performance, we follow~\newcite{DBLP:journals/corr/abs-2311-04850} and adopt three strategies, namely \textit{text transformation}, \textit{key information replacement}, and \textit{text concatenation} for fiction data augmentation. 

\paragraph{Text Transformation.} We utilize LLMs to facilitate mutual translation of fictions between Chinese and English, whereby the original Chinese (English) novels are rendered into English (Chinese). In accordance, the input and output for each task are also translated.

\paragraph{Key Information Replacement.} We employ LLMs to extract key information from a chapter or section, such as names, places, and times. We then generate corresponding texts to replace these elements, resulting in a collection of $\langle$original text - replacement text$\rangle$ pairs, which are subsequently used for content substitution throughout the entire text and tasks.

\paragraph{Text Concatenation.} We insert a short story into the original fiction as one of its chapters, and use this template to bridge: \textit{Now, let’s pause the current story narration and turn to a new story}[New Story]\textit{The story is over, let’s get back to the original fiction}. Then, we merge the data in four tasks of this short story with the original fiction.

Through above three strategies, we construct \textbf{Fiction-T} (Translated), \textbf{Fiction-R} (Replaced), and \textbf{Fiction-C} (Concatenated). These three datasets can ensure that the model must fully comprehend the entire text in order to accomplish tasks, rather than being able to complete tasks by recalling the content of the training phase.

\begin{table}[!t]
\centering
\resizebox{1\columnwidth}{!}{
\begin{tabular}{l|ccccc}
\toprule[1.3pt]
\multicolumn{1}{l|}{\multirow{2}{*}{\textbf{Models}}} & \multicolumn{5}{c}{\textbf{Tasks}}                                                                                        \\  \cmidrule{2-6}
\multicolumn{1}{l|}{}                       & \textit{C-S} & \textit{QA} & \textit{B-S} & \textit{Re-A} & \textit{N-G} \\ \cmidrule{1-6}
YaRN-Mistral-7B                      & 5.20                  & 5.00               & 5.11                     & 5.00                  & 5.22            \\
InternLM2-Chat-7B                   & 5.11                  & 5.00               & 5.58                     & 5.00                  & 5.37            \\
InternLM2-Chat-20B                   & 2.87                  & 1.23               & 3.37                     & 3.20                  & 1.41            \\
Moonshot-V1-128K                           & {\ul 2.19}            & \textbf{1.03}      & {\ul 1.21}               & 1.31                  & {\ul 1.02}      \\
GLM-4                               & 2.29                  & 1.11               & 1.25                     & \textbf{1.16}         & 1.19            \\
GPT-4-Turbo                        & \textbf{2.17}         & {\ul 1.06}         & \textbf{1.10}            & {\ul 1.23}            & \textbf{1.00}  \\ \bottomrule[1.3pt]
\end{tabular}
}
\caption{Human evaluation results of six LLMs on five tasks of Chinese Fiction Reading. The abbreviations in the table are consistent with those in the preceding tables. Lower scores indicate better performance.\label{human_evaluation} }
\end{table}

\subsection{Implementation Details}
We select GPT-4-Turbo~\cite{achiam2023gpt} to help us construct \benchname. GPT-4 currently stands as the highest-performing LLMs, characterized by a 128k context window along with superior memory, reasoning, and generation capabilities. The prompts and input templates used throughout the construction process are available in our GitHub repository due to space limit.

%% file: sections/Exp.tex
\section{Experimental Settings}


\begin{table*}[!t]
\centering
\resizebox{1.8\columnwidth}{!}{
\begin{tabular}{l|cc|cc|cc|c}
\toprule[1.3pt]
\multirow{2}{*}{\textbf{Models}} & \multicolumn{2}{c|}{\textbf{MR}}         & \multicolumn{2}{c|}{\textbf{DU}}   & \multicolumn{2}{c|}{\textbf{OU}} & \textbf{TG}      \\ \cmidrule(lr){2-3} \cmidrule(lr){4-5} \cmidrule(lr){6-7} \cmidrule(lr){8-8}
                                 & \textit{LE-L} & \textit{LE-R} & \textit{Def-QA} & \textit{Num-QA} & \textit{LE-C} & \textit{MCQA}  & \textit{Case-Adj} \\ \cmidrule{1-8}
YaRN-Mistral-7B-128K             & \textless{}1       & 11.29              & 8.62            & \textless{}1    & 3.36           & \textless{}1   & \textless{}1     \\
InternLM2-Chat-7B-200K           & \textless{}1       & 2.61               & 3.52            & \textless{}1    & \textless{}1   & \textless{}1   & \textless{}1     \\
InternLM2-Chat-20B-200K          & 5.41               & 22.60              & 40.57           & 58.03           & 11.76          & 44.23          & 41.05            \\
Moonshot-V1-128K                 & \textbf{32.61}     & \textbf{88.83}     & \textbf{48.08}  & {\ul 63.85}     & 28.10          & {\ul 63.11}    & {\ul 47.40}      \\
GLM-4-200K                      & {\ul 16.97}        & {\ul 72.76}        & {\ul 43.17}     & \textbf{67.63}  & \textbf{31.14} & 53.56          & 47.31            \\
GPT-4-Turbo-128K                & 13.41              & 63.48              & 40.26           & 62.50           & {\ul 29.51}    & \textbf{63.24} & \textbf{48.89}  \\  \bottomrule[1.3pt]
\end{tabular}
}
\caption{Results (\%) of six LLMs on Law Reading. \textbf{LE-L}, \textbf{LE-R}, \textbf{Def-QA}, \textbf{Num-QA}, \textbf{LE-C}, \textbf{MCQA} and \textbf{Case-Adj} represent \textit{Legal Entry Location}, \textit{Legal Entry Retrieval}, \textit{Legal Definition QA}, \textit{Legal Number QA}, \textit{Legal Entry Counting}, \textit{ Multiple Choice QA} and \textit{Case Adjudication}, respectively. Rest settings remain the same as in the previous tables. \label{results_law_reading}}
\end{table*}

\begin{table*}[!t]
\centering
\resizebox{1.8\columnwidth}{!}{
\begin{tabular}{l|cc|cc|cc|c}
\toprule[1.3pt]
\multirow{2}{*}{\textbf{Models}} & \multicolumn{2}{c|}{\textbf{MR}}                                                 & \multicolumn{2}{c|}{\textbf{DU}}                                           & \multicolumn{2}{c|}{\textbf{OU}}                                        & \multicolumn{1}{c}{\textbf{TG}}      \\  \cmidrule(lr){2-3} \cmidrule(lr){4-5} \cmidrule(lr){6-7} \cmidrule(lr){8-8}
                                 & \textit{LE-L} & \textit{LE-R} & \textit{Def-QA} & \textit{Num-QA} & \textit{LE-C} & \textit{MCQA}  & \textit{Case-Adj} \\  \cmidrule{1-8}
InternLM2-Chat-20B-200K                   & \textbf{5.41}                          & \textbf{22.60}                         & \textbf{40.57}                      & \textbf{58.03}                      & \textbf{11.76}                     & \textbf{44.23}                    & \textbf{41.05}                       \\
\textit{w/} Sentence-Transformers            & \textless{}1                           & 16.54                                  & 11.59                               & 11.22                               & 4.92                               & 39.92                             & 31.16                                \\
\textit{w/} LLM-Embedder                     & 1.86                                   & 21.68                                  & 11.97                               & 19.98                               & 2.46                               & 42.59                             & 38.83                                \\
\textit{w/} Contriever                       & \textless{}1                           & 16.73                                  & 10.23                               & 5.44                                & 4.10                               & 40.23                             & 37.79     \\  \bottomrule[1.3pt]                            
\end{tabular}
}
\caption{Results (\%) of InternLM2-Chat-20B-200K using different embedding models on Law Reading. \textit{w/} represents \textit{with}. The best performance over of each subtask is in \textbf{bold}.\label{results_rag}}
\end{table*}

\subsection{Generative Large Language Models}
We introduce current LLMs with context window size \textbf{more than 100k} evaluated in our experiments. Models such as LLama2~\cite{DBLP:journals/corr/abs-2307-09288} and ChatGLM2~\cite{DBLP:conf/iclr/ZengLDWL0YXZXTM23} have context window size significantly shorter than the average text length of \benchname, resulting in an excessive need to truncate texts, which leads to suboptimal performance. Consequently, we do not evaluate the effectiveness of these models.

\paragraph{GPT-4-Turbo} Developed by OpenAI, GPT-4-Turbo represents the pinnacle of current advancements, demonstrating exceptional reasoning and instruction-following capacities. It is distinguished by its extensive context window of 128K tokens. We employ this model via API\footnote{https://chat.openai.com/}.

\paragraph{GLM-4} GLM-4 is the latest model developed by Zhipu AI. Compared to ChatGLM2, it boasts more powerful question-answering and text generation capabilities, capable of processing up to 200,000 tokens. We employ this model via API\footnote{https://open.bigmodel.cn/}.

\paragraph{Moonshot-V1} Developed by Moonshot AI, Moonshot-V1 boasts exceptional performance in processing extremely-long text inputs of up to 128K tokens. We employ this model (\textbf{released before 2024.1.23}) via API\footnote{https://www.moonshot.cn/}.

\paragraph{InternLM2-Chat} Equipped with a 200k context window, InternLM2 exhibits comprehensive enhancements across all functionalities when juxtaposed with the previous generation model. We employ InternLM2-Chat-7B-200k and InternLM2-Chat-20B-200k.

\paragraph{YaRN-Mistral} The computationally efficient length extrapolation technology YaRN makes it possible to expand LLM’s context window size while conserving resources. We leverage YaRN-Mistral-7B-128k.

\subsection{Retrieval-Augmented Generation Methods}
One type of methods to handle long texts with small context window size in LLMs is Retrieval-Augmented Generation (RAG)~\cite{DBLP:journals/corr/abs-2202-01110}. Given a long context, we first splits it into chunks. Then, using a specific retriever, we compute the embedding of the text chunks and query. Only the top-N chunks, based on the cosine similarity of their embeddings to the query embedding, are concatenated. These top-N chunks along with the query are then fed into the model to produce an answer. We test this technique’s impact on LLMs evaluation results, to see if the model could complete \benchname tasks by retrieving certain fixed chunks. We employ LangChain\footnote{\href{https://python.langchain.com/docs/get\_started/introduction}{https://python.langchain.com/docs/get\_started/introduction}} and three retrievers: Sentence-Transformers~\cite{reimers-2020-multilingual-sentence-bert}, LLM-Embedder~\cite{llm_embedder}, and Contriver~\cite{DBLP:journals/tmlr/IzacardCHRBJG22}. We set the chunk size to 500 and N=5.

\begin{table*}[!t]
\centering
\resizebox{1.8\columnwidth}{!}{
\begin{tabular}{l|ccc|ccc}    
\toprule[1.3pt]
\multirow{2}{*}{\textbf{Models}} & \multicolumn{3}{c}{\textbf{Paper Reading}}     & \multicolumn{3}{|c}{\textbf{Law Reading}}           \\ \cmidrule{2-7}
                                 & \textit{Content-R} & \textit{Sec-Sum} & \textit{T-Explain} & \textit{LE-L}  & \textit{Def-QA} & \textit{Num-QA} \\ \cmidrule{1-7}
\rowcolor{gray!50}
YaRN-Mistral-7B-128K             & \textless{}1 & 10.19            & 15.86        & \textless{}1   & 8.62            & \textless{}1    \\
\textit{w/} ICL    & \textless{}1 & 9.81           & 14.30      & \textless{}1 & 7.79          &  \textless{}1  \\ \cmidrule{1-7}
\rowcolor{gray!50}
InternLM2-Chat-20B-200K          & 25.84        & 24.91            & 30.27        & 5.41           & 40.57           & 58.03           \\
\textit{w/} ICL    &  31.89      &  33.67          &  38.50      &  6.76         & 39.90         &  58.82         \\ \cmidrule{1-7}
\rowcolor{gray!50}
GPT-4-Turbo-128K                 & 45.28        & 51.57            & 55.91        & 13.41          & 40.26           & 62.50           \\
\textit{w/} ICL   &  46.77      &  50.89          &  56.12      &  14.81        &  40.88         &  61.58      \\ \bottomrule[1.3pt]   
\end{tabular}
}
\caption{Results (\%) of three LLMs using zero-shot learning and few-shot learning on several tasks of Paper Reading and Law Reading. The data in the \textbf{gray} section is derived from the previous tables.\label{fsl}}
\end{table*}

\begin{figure}[t]
	\centering
	\includegraphics[width=\linewidth]{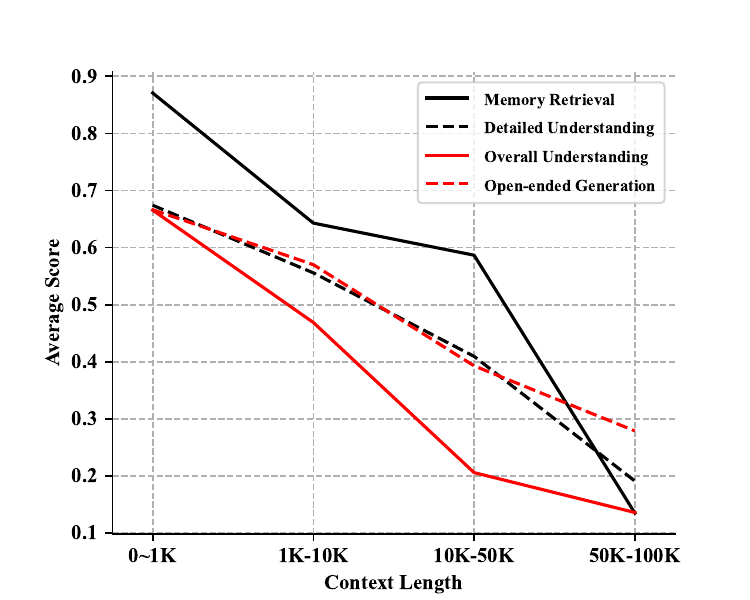}
	\caption{Average score (\%) of four tasks under different context length on Law Reading.}
	\label{result_length}
    \vspace{0mm}
\end{figure}

\subsection{Automatic Evaluation Metrics}
For tasks with fixed answers, such as Content Location in Fiction Reading, we adopt \textbf{Accuracy} as an intuitive measure to demonstrate the model’s performance. For MCQA, we utilize \textbf{F1-Score} to objectively evaluate the model’s capability to accurately answer all the correct options. For summary tasks, we select \textbf{Rouge-L} to reflect whether the model can correctly identify key information in a document. For generative tasks, we employ \textbf{BLEU} to measure the congruence between the generated content by model and the reference content. For Rating Score subtask, we choose MAE to calculate the average absolute difference between predicted and true scores. Details can be found in Table \ref{benchmark}.

\subsection{Inference Settings}
We conduct the evaluation in a zero-shot setting. The input templates we use during inference can be found in Appendix~\ref{template}. When the input length exceeds the context window size of LLMs, we truncate the input sequence from the middle, as the front and end of the sequence may contain crucial information such as instructions or questions. For models that are API-callable, we follow the original settings provided in the sample code of these models. For locally deployed models, we select the decoding parameters as follows: Temperature=$0.2$, Top-K=$40$, Top-P=$0.9$, Repetition Penalty=$1.02$.

\begin{table*}[!t]

\resizebox{2.1\columnwidth}{!}{
\begin{tabular}{l|cc|cc|cccccc|ccc}
\toprule[1.3pt]
\multirow{2}{*}{\textbf{Scenarios}}  & \multicolumn{2}{c|}{\textbf{MR}}       & \multicolumn{2}{c|}{\textbf{DU}} & \multicolumn{6}{c|}{\textbf{OU}}                                                                     & \multicolumn{3}{c}{\textbf{TG}}    \\ \cmidrule(lr){2-3} \cmidrule(lr){4-5} \cmidrule(lr){6-11} \cmidrule(lr){12-14}  
 & \textit{C-L}  & \textit{C-R}          & \textit{C-S}    & \textit{QA}   & \textit{C-Q}   & \textit{F-B}   & \textit{F-E}   & \textit{F-S}   & \textit{Ch-D}  & \textit{Ch-R}  & \textit{Ch-DG} & \textit{N-G}   & \textit{P-G}   \\ \cmidrule{1-14}
Fiction         & \textbf{6.85} & \textbf{\textless{}1} & \textbf{17.22}  & \textbf{9.82} & \textbf{53.33} & \textbf{15.58} & \textbf{18.61} & \textbf{17.29} & \textbf{21.98} & \textbf{28.92} & 11.65          & \textbf{16.67} & \textbf{10.09} \\
Fiction-T       & 6.54          & \textless{}1          & 12.28           & 5.05          & 52.16          & 10.21          & 10.80          & 6.67           & 2.28           & 13.89          & 12.36          & 11.89          & 5.01           \\
Fiction-R       & 6.76          & \textless{}1          & 5.11            & 6.48          & 53.33          & 8.04           & 11.72          & 4.96           & 3.33           & 17.67          & 12.12          & 11.84          & 5.78           \\
Fiction-C       & 6.28          & \textless{}1          & 5.23            & 3.39          & 53.33          & 7.65           & 4.46           & 13.41          & 2.49           & 15.56          & \textbf{13.79} & 12.68          & 7.91      \\ \bottomrule[1.3pt]     
\end{tabular}
}
\caption{Results (\%) of six LLMs on Fiction, Fiction-T, Fiction-R, and Fiction-C. \label{results_ablation}}
\end{table*}

\section{Results and Analysis}
\subsection{Long Texts Processing}
The results pertaining to three scenarios are delineated in Table \ref{results_fiction_cn} and \ref{results_law_reading}. Due to space constraints, the remaining results are relegated to Appendix \ref{all}. The key findings from the experiments can be summarized below.

\textbf{The overall performance of all LLMs is notably unsatisfactory.} Regardless of whether they are open-source or closed-source, LLMs consistently score low across various metrics pertaining to the 27 subtasks, particularly in retrieval and counting tasks where human performance approaches 100\%. We hypothesize that these results are attributable to the use of sparse attention or length extrapolation techniques within the extended model context window, as well as the truncation operation employed when the input text is too long. 


\textbf{Closed-source models outperform open-source models.} The comparative performance analysis of three closed-source LLMs demonstrates a superior performance over their open-source counterparts. Furthermore, with 7B parameters, YaRN-Mistral and InternLM2-Chat-7B exhibit suboptimal performance across a majority of tasks, achieving scores below 1. This demonstrates the importance of the model's parameter size for effectively managing tasks in \benchname.

\textbf{LLMs have a preference for the language of the input text.} GLM-4 and Moonshot-V1 performs well on Chinese-language tasks (Law Reading and Fiction-CN), while GPT-4 performs well on English-language tasks (Paper Reading and Fiction-EN). We infer that this may be due to the different proportions of Chinese and English datasets used in the training process of these three models. This further indicates that the dataset is a particularly critical factor that affects model performance.

\textbf{GPT-4's performance on self-generated subtasks does not meet expectations.} In particular, for subtasks where the ground truth is established by GPT-4 itself, we meticulously assessed the model's efficacy. Contrary to our initial assumptions, GPT-4's scores on these tasks were lower than anticipated. Upon an in-depth analysis of the model-generated content, we hypothesized that the verbose nature of the text could have adversely affected GPT-4's understanding of the task descriptions, leading to a diminished output quality.

The findings and analyses presented above indicate that existing context window expansion technologies fall significantly short of reaching or approximating human-level performance. Addressing the issue of context dependency represents a critical area for potential breakthroughs and merits further exploration.
\subsection{Human Evaluation Results}
It has been correctly noted that numerous studies have exposed significant limitations in N-grams matching-based metrics, such as Rouge-L and BLEU\cite{DBLP:conf/iclr/ZhangKWWA20}. To address the shortcomings associated with these metrics, we engage volunteers to perform human evaluation on fiction reading. We utilize a consistent dataset of \textbf{100} inputs across several tasks. Volunteers are presented with outputs from all LLMs simultaneously. They are then asked to rank these outputs based on perceived quality. Our ranking system accommodates ties, with subsequent rankings adjusted to reflect these equivalences.\footnote{For example, the rankings might be represented as 1, 1, 3, 4, 5, 5.}
We present the average ranking of each model across all tasks, highlighting the following outcomes. 

As shown in Table~\ref{human_evaluation}, the models of 7B size consistently occupy the bottom two rankings across all evaluated tasks. Further case analysis demonstrates that their outputs are characterized by disorganization and incoherence, often devoid of logical structure or bordering on nonsensical. Meanwhile, the 20B InternLM2-Chat typically secures the fourth position, and the average rankings of the other three LLMs are exclusively accessible via API calls are closely contested.

\subsection{Performance of Retrieval-Augmented Generation Methods}
In this subsection, we assess the performance of InternLM2-Chat-20B-200K, which utilizes three distinct retrievers on Law Reading scenarios. Results illustrated in Table~\ref{results_rag}, indicate a uniform reduction in the model's performance across all subtasks following the adoption of RAG methods. Notably, the most substantial declines in performance were observed in the Definition QA and Number QA tasks. We postulate that these decreases may be due to the retrievers' failure to recall relevant segments of text. The results and subsequent analysis imply that effectively addressing the tasks in \benchname demands more than merely retrieving relevant documents.

\subsection{Assessment of In-context Learning Ability}
Previous analysis primarily focuses on the zero-shot setting. In this section, we assess LLMs' In-context Learning (ICL) capability on selected tasks. Due to constraints on input length, we opt to use \textbf{two} samples from identical long texts and tasks as prompts, assessing the performance of LLMs on the remaining random \textbf{200} data. 

Table~\ref{fsl} demonstrates significant enhancements primarily in summarization tasks. Through in-context learning, models are capable of generating outputs that closely align with the desired format, thus elevating their scores. Conversely, in tasks necessitating brief responses, models exhibit limited ability to leverage the prompts for noticeable improvement. Specifically, GPT-4-Turbo, despite its substantial parameter count, shows negligible performance shifts following in-context learning application.

\subsection{Impact of Context Length}
In this subsection, we explore the impact of context length on the performance of LLMs. Our evaluation focuses on the average performance of the InternLM2-Chat-20B across four tasks, using legal texts of varying lengths. Figure~\ref{result_length} illustrates that the model's performance significantly declines with longer texts, as evidenced by a steeper curve. This observation underscores the model's challenges in effectively managing the complexities of long text modeling.

\subsection{Impact of Data Contamination}
In this section, we assess the effectiveness of our data augmentation strategies in mitigating the impact of data contamination on model evaluation outcomes. We specifically examine the performance of the InternLM2-Chat-20B across different subsets of fiction data, namely Fiction, Fiction-T, Fiction-R, and Fiction-C, with the results detailed in Table~\ref{results_ablation}. The observed reduction in performance across almost all subtasks within the augmented dataset indicates that our data augmentation techniques can, to some extent, reduce the likelihood of biased evaluations.

%% file: sections/Conclusion.tex
\section{Conclusion}
In this paper, we present \benchname, a comprehensive benchmark for extremely long text understanding with long-range dependencies. \benchname consists of three scenarios, four tasks, and 27 subtasks, with an average length of over 100K words (English) and 200K characters (Chinese). We automatically construct the benchmark via LLMs, significantly reducing the cost of manually annotating the datasets. Furthermore, we mitigate data contamination risks through carefully designed techniques. Extensive experiments on \benchname yield insights into the capabilities of current LLMs for long text understanding. We also demonstrate that RAG methods are not suitable for \benchname as the benchmark requires a comprehensive understanding of the entire text to complete the tasks. Results and analyses indicate that \benchname is a valuable resource for advancing research in the comprehension of long texts.

\section*{Limitations}
The limitations of \benchname mainly come from the disadvantages of using LLMs. First of all, most of the large language models that work well are not open source or free. This makes it difficult to conduct batch experiments or daily use on it. Next, a small number of open-source models require a lot of GPU resources when used, which is a difficult problem for quite many researchers, such as students.

\section*{Ethics Statement}
We honor and support the ACL code of Ethics. Our bencmark \benchname aims to evaluate large language models' ability of long-text comprehension. The interaction and assistance process do not involve any bias towards to the participants. Following our thorough examination, we can confirm that our benchmark is free from any privacy or ethical concerns.

\section*{Acknowledgements}
We would like to extend our gratitude to Moonshot AI for their support in LLMs evaluation. This research is supported by the National Natural Science Foundation of China (No.62106105), the CCF-Baidu Open Fund (No.CCF-Baidu202307), the CCF-Zhipu AI Large Model Fund (No.CCF-Zhipu202315), the Scientific Research Starting Foundation of Nanjing University of Aeronautics and Astronautics (No.YQR21022), and the High Performance Computing Platform of Nanjing University of Aeronautics and Astronautics.

%% file: sections/A-Description.tex
\section{Task Descriptions}
\label{detailed description}
\begin{table*}[!t]

\resizebox{2.1\columnwidth}{!}{
\begin{tabular}{l|cc|cc|cccccc|ccc}
\toprule[1.3pt]
\multirow{2}{*}{\textbf{Models}} & \multicolumn{2}{c|}{\textbf{MR}}                 & \multicolumn{2}{c|}{\textbf{DU}}          & \multicolumn{6}{c|}{\textbf{OU}}                                                                                        & \multicolumn{3}{c}{\textbf{TG}}      \\ \cmidrule(lr){2-3} \cmidrule(lr){4-5} \cmidrule(lr){6-11} \cmidrule(lr){12-14}  
                                 & \textit{C-L} & \textit{C-R} & \textit{C-S} & \textit{QA} & \textit{C-C} & \textit{B-S} & \textit{E-E} & \textit{F-S} & \textit{Ch-D} & \textit{Re-A} & \textit{RP-C} & \textit{N-G} & \textit{P-G} \\ \cmidrule{1-14}
YaRN-Mistral-7B             & \textless{}1           & \textless{}1           & 6.64                 & 2.29           & 5.52               & 10.16              & 2.85              & 3.13                 & 10.09           & 8.52            & 4.36                 & 4.42            & 5.40            \\
InternLM2-C-7B           & \textless{}1           & \textless{}1           & 3.08                 & \textless{}1   & \textless{}1       & 7.73               & 5.15              & 4.57                 & 7.01            & 2.31            & 6.90                 & 4.23            & 21.88           \\
InternLM2-C-20B        & 18.85                  & 1.58                   & 17.60                & 35.43          & 56.01              & 17.47              & 29.81             & 25.04                & 19.97           & 20.73           & 53.14                & 29.79           & 44.81           \\
Moonshot-V1                   & {\ul 38.19}            & 33.56                  & \textbf{24.46}       & {\ul 34.14}    & \textbf{88.89}     & \textbf{30.30}     & {\ul 38.79}       & {\ul 39.16}          & {\ul 28.45}     & 25.46           & 37.10                & {\ul 61.76}     & {\ul 62.47}     \\
GLM-4                    & 26.68                  & {\ul 34.60}            & 18.06                & 32.86          & 66.67              & 28.75              & 34.46             & 24.30                & 25.24           & \textbf{27.56}  & {\ul 39.20}          & 35.07           & 53.12           \\
GPT-4-Turbo             & \textbf{55.46}         & \textbf{42.70}         & {\ul 19.76}          & \textbf{50.81} & {\ul 77.50}        & {\ul 29.30}        & \textbf{44.20}    & \textbf{42.57}       & \textbf{30.87}  & {\ul 27.16}     & \textbf{66.71}       & \textbf{74.59}  & \textbf{67.80} \\ \bottomrule[1.3pt] 
\end{tabular}
}
\caption{Results (\%) of six LLMs on English Fiction Reading.  \label{results_fiction_EN}}
\end{table*}

\begin{table*}[t]
\centering
\resizebox{1.9\columnwidth}{!}{
\begin{tabular}{l|c|cc|cc|cc}
\toprule[1.3pt]
\multirow{2}{*}{\textbf{Models}} 
                        & \multicolumn{1}{c|}{\textbf{MR}} & \multicolumn{2}{c|}{\textbf{DU}}   & \multicolumn{2}{c|}{\textbf{OU}} & \multicolumn{2}{c}{\textbf{TG}} \\ \cmidrule(lr){2-2} \cmidrule(lr){3-4} \cmidrule(lr){5-6} \cmidrule(lr){7-8}
                        & \textit{C-R}            & \textit{Sec-Sum} & \textit{T-E} & \textit{Paper-C}        & \textit{Paper-Sum} & \textit{P-Review}         & \textit{R-Score}↓         \\ \cmidrule{1-8}
YaRN-Mistral-7B-128K    & \textless{}1               & 10.19                 & 15.86   & 11.69        & 5.04          & 33.23          & None          \\
InternLM2-Chat-7B-200K  & \textless{}1               & 6.82                  & 5.04    & \textless{}1 & 7.31          & 39.80          & None          \\
InternLM2-Chat-20B-200K & 25.84                      & 24.91                 & 30.27   & 33.37        & 34.41         & 45.11          & \underline{2.30}          \\
Moonshot-V1-128K          & \underline{31.02}                      & \underline{45.78}                 & 31.43   & 44.44        & 36.68         & \textbf{66.04}          & 4.39          \\
GLM-4-200K              & 25.76                      & 29.66                 & \underline{33.40}   & \underline{47.62}        & \underline{36.91}         & 55.62          & \textbf{2.23}         \\
GPT-4-Turbo-200K       & \textbf{45.28}        & \textbf{51.57}                 & \textbf{55.91}   & \textbf{55.56 }       & \textbf{45.91}         & \underline{62.12}          & 2.63  \\ \bottomrule[1.3pt]
\end{tabular}
}
\caption{Results (\%) of six LLMs on Paper Reading. \textbf{Sec-Sum}, \textbf{T-E}, \textbf{Paper-C}, \textbf{Paper-Sum}, \textbf{P-Review}, and \textbf{R-Score} represent \textit{Section Summarization}, \textit{Terminology Explanation}, \textit{Paper Counting}, \textit{Paper Summarization}, \textit{Paper Review}, and \textit{Rating Score} respectively. \textbf{None} signifies the model's inability to generate a rating score, thus rendering it incapable of fulfilling the requirements of this subtask. \label{results_paper_reading}}
\end{table*}

In this section, we provide detailed descriptions of the input and output content of 27 subtasks. Please note that the input includes a long text and an instruction. We only describe the instruction.
\subsection{Fiction Reading}
\paragraph{Content Location} Given the content of the fiction, the model outputs the location.

\paragraph{Content Retrieval} Given a location, the model outputs the corresponding fiction content.

\paragraph{Chapter Summarization} Given a chapter number of the fiction, the model summarizes the corresponding chapter.

\paragraph{Question Answering} Give a detailed question about the fiction, the model outputs the answer.
 
\paragraph{Chapter Counting} The model outputs the quantity of the fiction.

\paragraph{Background Summarization} The model outputs the time background, place background, and social and cultural background of the fiction.

\paragraph{Event Extraction} The model outputs the main events of the fiction in chronological order.

\paragraph{Fiction Summarization} The model summarizes the whole fiction.

\paragraph{Character Description} The model outputs the description of the character in the fiction, including personality traits and personal experiences.

\paragraph{Relationship Analysis} The model outputs the relationship between two characters.

\paragraph{Role-play Conversation} Given a question, the model needs to assume the role of a character from the fiction to provide an answer.

\paragraph{News Generation} The model assume a news reporter within the world of the fiction, and reports on the final event involving the protagonist's team, including the background of the event, the actions of the protagonist, the outcome, and the impact of the event.

\paragraph{Poem Generation} The model writes a poem based on the core theme, key plot, important characters and specific context of the fiction.

\subsection{Paper Reading}

\paragraph{Content Retrieval} Given a location, the model outputs the corresponding paper content, such as title, authors.

\paragraph{Section Summarization} Given a section number of the paper, the model summarizes the corresponding section.

\paragraph{Terminology Explanation} Given an scientific noun in the paper, the model outputs its explanation.

\paragraph{Paper Counting} The model output the quantity of titles, authors, references, tables, figures, etc. of the paper.

\paragraph{Paper Summarization} The model summarizes the whole paper.

\paragraph{Paper Review} The model assumes the role of a peer reviewer for an academic journal, and outputs a review of the paper, including: strengths and weaknesses.

\paragraph{Rating Score} The model assumes the role of a peer reviewer for an academic journal, and outputs a rating score of the paper from 0 to 10.

\subsection{Law Reading}

\paragraph{Legal Entry Location} Given the content of the law, the model outputs its corresponding index.
\paragraph{Legal Entry Retrieval} Given a locating of a legal entry, the mode outputs its content.
\paragraph{Legal Definition QA} Given a question about the law's definitions, the model outputs the answer.
\paragraph{Legal Number QA} Given questions about the numbers in law, the model outputs the answer.
\paragraph{Legal Entry Counting} The model outputs the quantity of legal entries in this law.
\paragraph{Multiple Choices QA} Given a question with multiple choices, the model outputs the answer. 
\paragraph{Case Adjudication} Given a legal case, the model outputs the verdict.

%% file: sections/A-template.tex
\section{Evaluation Input Templates}
\label{template}
For all texts and corresponding questions in \benchname, we use the following template: \textit{Please read the following text, and answer related question:} [text] \textit{Question:} [question] \textit{Directly output your answer without any additional analysis or explanation.}

%% file: sections/A-All_Results.tex
\section{Results on English Fiction Reading and Paper Reading}
\label{all}

We show the remaining results of six LLMs on English Fiction Reading and Paper Reading in Table~\ref{results_fiction_EN} and Table~\ref{results_paper_reading}.